\documentclass[lettersize,journal]{IEEEtran}
\usepackage{amsmath,amsfonts}
\usepackage{algorithmic}
\usepackage{algorithm}
\usepackage{array}
\usepackage[caption=false,font=normalsize,labelfont=sf,textfont=sf]{subfig}
\usepackage{textcomp}
\usepackage{stfloats}
\usepackage{url}
\usepackage{verbatim}
\usepackage{graphicx}
\usepackage{cite}
\usepackage{bbm}
\usepackage{diagbox} 
\usepackage{booktabs, multirow, multicol}
\usepackage{makecell}
\usepackage{pifont}
\usepackage{xcolor} 
\usepackage{hyperref}
\hypersetup{hidelinks}
\DeclareUnicodeCharacter{FF0C}{}

\begin{document}

\title{Boosting Semi-Supervised Object Detection in Remote Sensing Images with Active Teaching}

\author{Boxuan Zhang, Zengmao Wang,~\IEEEmembership{Member,~IEEE} and Bo Du,~\IEEEmembership{Senior Member,~IEEE}
        % <-this % stops a space
        
% \thanks{}% <-this % stops a space
% \thanks{}
\thanks{ This work was supported in part by the National Natural Science Foundation of China under Grants 62271357, the Natural Science Foundation of Hubei Province under Grants 2023BAB072, the Fundamental Research Funds for the Central Universities under Grants 2042023kf0134.\textit{(Corresponding author:
Zengmao Wang.)}

Boxuan Zhang is with the School of Computer Science, Wuhan University, Wuhan 430072, China(e-mail: \href{mailto:zhangboxuan1005@gmail.com}{zhangboxuan1005@gmail.com}).

Zengmao Wang and Bo Du are with the National Engineering Research Center for Multimedia
Software, School of Computer Science, Institute of Artificial Intelligence, and the Hubei Key Laboratory of Multimedia and Network Communication Engineering, Wuhan University, Wuhan 430072, China (e-mail: \href{mailto:wangzengmao@whu.edu.cn}{wangzengmao@whu.edu.cn}; \href{mailto:gunspace@163.com}{gunspace@163.com}).

% Color versions of one or more of the figures in this letter are available
% online at http://ieeexplore.ieee.org.
}}

% The paper headers
\markboth{Journal of \LaTeX\ Class Files,~Vol.~14, No.~8, August~2021}%
{Shell \MakeLowercase{\textit{et al.}}: A Sample Article Using IEEEtran.cls for IEEE Journals}

% \IEEEpubid{0000--0000/00\$00.00~\copyright~2021 IEEE}
% Remember, if you use this you must call \IEEEpubidadjcol in the second
% column for its text to clear the IEEEpubid mark.

\maketitle

\begin{abstract}

The lack of object-level annotations poses a significant challenge for object detection in remote sensing images. To address this issue, active learning and semi-supervised learning techniques have been proposed to enhance the quality and quantity of annotations. Active learning focuses on selecting the most informative samples for annotation, while semi-supervised learning leverages the knowledge from unlabeled samples. In this paper, we propose a novel active learning method to boost semi-supervised object detection for remote sensing images with a teacher-student network, called SSOD-AT. The proposed method incorporates a RoI Comparison module (RoICM) to generate high-confidence pseudo-labels for Regions of Interest (RoIs). Meanwhile, the RoICM is utilized to identify the top-K uncertain images. To reduce redundancy in the top-K uncertain images for human labeling, a diversity criterion is introduced based on object-level prototypes of different categories using both labeled and pseudo-labeled images. Extensive experiments on DOTA and DIOR two popular datasets demonstrate that our proposed method outperforms state-of-the-art methods for object detection in remote sensing images. Compared with the best performance in the SOTA methods, the proposed method achieves 1\% improvement at most cases in the whole active learning.

\end{abstract}


\section{Introduction}

\IEEEPARstart{O}{bject} detection in remote sensing images is a crucial task to identify the objects with their locations \cite{intro-1}. In recent years, deep learning has shown promising results in object detection for remote sensing images. However, the success of deep learning approaches heavily relies on large-scale datasets with accurately labeled data, which are typically annotated by human experts. Unlike image classification, object detection requires object-level labels that include both bounding box coordinates and object categories. This annotation process is more laborious and time-consuming. Additionally, remote sensing images often contain objects with various orientations and scales \cite{intro-2}, \cite{intro-3}, \cite{intro-4}, \cite{intro-5}, \cite{intro-6}, which further complicates the labeling process. Hence, although the collection of remote sensing images are to be faster and easier with the advancements of remote sensing technology, the availability of labeled images for object detection is usually limited \cite{intro-7}, \cite{intro-8}, \cite{intro-9}.

Semi-supervised learning (SSL) and active learning (AL) are two promising techniques in machine learning to address the problem with limited labeled images. 
SSL usually attempts to exploit the unlabeled data with a limited amount of labeled data by assuming the consistency between the feature distribution of unlabeled data and labeled data. 
% SSL with pseudo-labeling \cite{pseudo-label} has achieved satisfactory performance in classification \cite{intro-semi-2}, \cite{r-s-meanteacher}, \cite{r-s-fixmatch}. 
For the semi-supervised object detection(SSOD) method, most of them are developed based on a teacher-student network, which introduces a secondary model (teacher) to guide the training of the primary model (student). The teacher network uses weakly augmented labeled data to generate high-quality pseudo-labels for the student network \cite{intro-ssod-aug-1}. The student network then is optimized by using the pseudo-labels as supervised information. 

Different from SSL, AL generally focuses on the labeled data, which are mainly collected by selecting the most informative samples from the unlabeled data for human experts labeling\cite{intro-active-1}. The query criterion is the core technique in AL methods, i.e. uncertainty and diversity\cite{intro-al-3}, \cite{intro-al-4}. Yuan et.al\cite{r-a-multi-instance} proposes MI-AOD, an instance-level uncertainty-based method which highlights the informative instances while filtering out noisy ones to select the most informative images for detector training. CALD\cite{r-a-cald} not only gauges individual information for sample selection but also leverages mutual information to alleviate unbalanced class distribution, thus ensuring the diversity of selected samples. MOL\cite{mol} introduces a temporal consistency-based instance selection strategy for discovering reliable foreground objects, reducing the risk of background interference.

 % Therefore, it is natural to consider combining AL and SSL to enhance the performance of object detection when dealing with limited labeled Remote Sensing Images (RSIs). In fact, several methods have been developed for Semi-Supervised Object Detection (SSOD) with AL in natural images. For instance, Lv et al. proposed a novel semi-supervised active salient object detection (SOD) method that utilized a salient encoder-decoder with an adversarial discriminator to select the most representative data. Mi et al. \cite{activeteacher} introduced the concept of "Active Teacher," which involves data initialization through AL using a teacher-student-based SSOD approach. However, these methods may not be directly applicable to remote sensing images due to their higher resolution and the presence of more complex objects.

We note that SSL usually relies on the labeled data to exploit the unlabeled data for task learning, while AL aims to label the most informative samples from the unlabeled data with a query criterion. Therefore, it is a natural consideration to combine AL and SSL together to improve the performance of object detection with limited labeled RSIs. In fact, many methods have been developed for SSOD with AL in natural images. Lv et al. \cite{sod} proposed a novel semi-supervised active salient object detection (SOD) method that utilized a salient encoder-decoder with an adversarial discriminator to select the most representative data. Mi et al. \cite{activeteacher} introduced the method "Active Teacher", which involves data initialization through AL using a teacher-student-based SSOD approach. However, these methods usually ignore the redundancy in the RoIs, resulting in that more images are selected to improve the performance for SSOD.

%Lv et.al proposed a novel semi-supervised active salient object detection(SOD) method which designed a salient encoder-decoder with adversarial discriminator to select the most representative data. Mi et al\cite{activeteacher} presented the first attempt of studying data initialization by AL with teacher-student based SSOD, called \textit{Active Teacher}. However, these methods may not be suitable for remote  sensing images because these images are often in high resolution and contain more complicated objects.

In this paper, we propose a method Semi-Supervised Object Detection with Active Teaching, termed as SSOD-AT, for object detection in remote sensing images using a teacher-student network. Remote sensing images often contain objects with large scale variations and high density, which is easy to produce noisy RoIs. To mitigate the impact of these noisy RoIs, we introduce a RoI Comparison Module (RoICM) that compares the RoIs generated by the teacher network and the student network. In our method, when the RoIs exhibit consistent predictions between the teacher and student networks, they are assigned pseudo-labels for semi-supervised training. Conversely, for RoIs with divergent predictions, we utilize them to calculate the uncertainty of the image based on the teacher network's predictions. To remove the redundancy from the queried images, a diversity is designed by introducing a global class prototype to ensure the diverse between the current query image and selected images. Finally, we integrate the uncertainty and diversity as the query score to select the most valuable images for human labeling. The main contributions of this article can be summarized as follows:

\begin{itemize}
\item{ A novel method to boost semi-supervised object detection with active learning is proposed for remote sensing images based on the teacher-student network. The proposed method can provide both confident pseudo-labels and informative images.}
\item{A RoI comparison module(RoICM) is introduced by comparing the RoIs generated by teacher and student network. It can effectively alleviate the influence of noisy RoIs for semi-supervised learning and improve the ability of active learning to evaluates the uncertainty of images.}
\item{The proposed method further incorporates the global class prototype for the diversity of selected images. The combination of the two sampling strategies maximizes the effectiveness of AL process.}
\end{itemize}

\section{Methodology}
\begin{figure*}[htb]
	\centering
	\includegraphics[width=15cm]{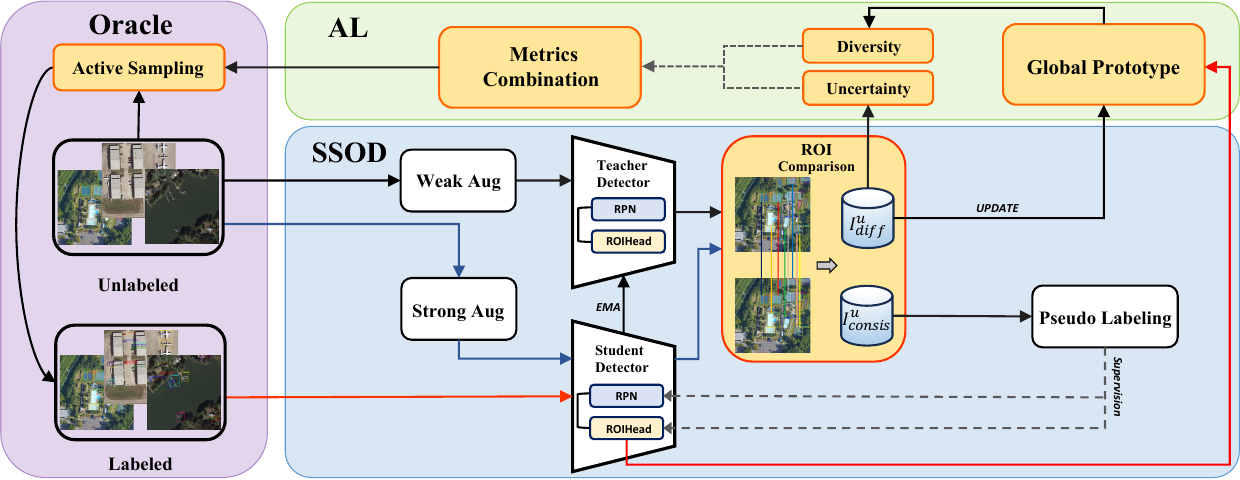}
	\caption{Overview of SSOD-AT framework for three stages. \textbf{\textit{Semi-Supervised Object Detection(SSOD)}}:  Using limited label set to initialize the parameters of Teacher-Student framework.
 % , where Teacher is responsible for generating pseudo-labels to Student while Student is trained with both ground-truth and pseudo-labels. 
    \textbf{\textit{Active Learning(AL)}}: Select the top-N valuable samples for labeling.
    % , which are based on RoI Comparison and Category Prototype, respectively. 
    \textbf{\textit{Label set Augmentation(Oracle)}}: Using the active selected samples to augment the label set. Repeat the preceding procedures to train the Teacher-Student framework.
    }
	\label{framework}
\end{figure*}
The proposed SSOD-AT method is shown in Fig.1. We introduce an iterative strategy to train Teacher-Student network with active learning for semi-supervised object detection. In SSOD-AT, a RoI Comparison Module is introduced by producing confident pseudo-labels for the ROIs and providing candidate ROIs for active learning to measure the uncertainty of the images. Meanwhile, a diversity criterion is designed based on the \textit{Global Class Prototype} to remove the redundancy with the queried images. In this section, we will introduce the details of the \textit{RoI Comparison Module(RoICM)} and the active sampling strategy.

%First, we initialize a small subset of labeled images as the label set and introduce a RoI Comparison Module to alleviate the noisy RoIs. After each round of semi-supervised training, we propose a novel active sampling strategy to select the most valuable images for annotation and update the label set. The strategy includes an uncertainty selection model assisted by the \textit{RoI Comparison Module(RoICM)} and a diversity selection model based on the \textit{Global Class Prototype}.
\subsection{RoI Comparison Module(RoICM) for Uncertainty selection}
Given an unlabeled image $x_{i}^u$,  we first input it into both the Teacher and Student networks to obtain their respective predictions of RoIs. These RoIs are then fed into the \textit{RoICM} for comparison. \textit{RoICM} employs a set of comparison rules that ensure the highest degree of accuracy and precision in the comparison process: If the RoIs predicted by both the Teacher and Student networks have \textit{consistent} classes, we add $x_{i}^u$ to the set of images with consistent class predictions $I_{consis}^{u}$. We deem the RoIs of $x_{i}^u \in I_{consis}^{u}$ predicted by the Teacher network to be reliable and proceed to pseudo-label the image, allowing to feed it into the Student network for the next stage of training. If the RoIs predicted by both the Teacher and Student networks have \textit{different} classes, we add $x_{i}^u$ to the set of images with different class predictions $I_{diff}^{u}$. This decision is predicated on the fact that $x_{i}^u \in I_{diff}^{u}$ is considered to have a higher level of uncertainty for the detection network, thereby increasing its overall annotation value. Thus, they should be included in the active learning selection sequence for human labeling.

Assisted by \textit{RoICM}, for $x_{i}^u \in I_{consis}^{u}$, we calculate the \textit{KL divergence} based on the predicted class distributions by the Teacher and Student networks:
\begin{equation}
	\label{kl}
	D_{k l}=\frac{1}{n_{b}^{i}} \sum_{j=1}^{n_{b}^{i}} \sum_{k=1}^{N_{c}} p_{t}^{i}\left(b_{j}, c_{k}\right) \frac{\log p_{t}^{i}\left(b_{j}, c_{k}\right)}{\log p_{s}^{i}\left(b_{j}, c_{k}\right)}
\end{equation}
where $n_{b}^{i}$ represents the number of proposal bounding boxes generated by the Teacher network after NMS and confidence threshold filtering, $N_{c}$ is the number of instance categories and $p^{i}\left(b_{j}, c_{k}\right)$ is the prediction probability of the k-th category by the network. $b^{'}_{j}$ and $b_{j}$ are the bounding boxes predicted by student and teacher network respectively. A lower $D_{k l}$ indicates more reliable pseudo-labels with a higher weight in the total loss as follows:
\begin{equation}
	\label{total_loss}
	\mathcal{L}_{det}^{ts}\left( \mathcal{D}_{l},  \mathcal{D}_{u}\right) = \mathcal{L} _{det}^{sup}\left( \mathcal{D}_{l} \right) + \exp(-D_{k l})\cdot \lambda _{u}\cdot \mathcal{L} _{det}^{unsup}\left( \mathcal{D}_{u} \right) 
\end{equation}

Instead, the level of uncertainty for $x_{i}^u \in I_{diff}^{u}$ is calculated by  the predicted category distribution of the Teacher network as follows:
\begin{equation}
	\label{unc}
	\cdot S_{u n c}^{i}=-\frac{1}{n_{b}^{i}} \sum_{j=1}^{n_{b}^{i}} \sum_{k=1}^{N_{c}} p_{t}^{i}\left(b_{j}, c_{k}\right) \log p_{t}^{i}\left(b_{j}, c_{k}\right)
\end{equation}

\subsection{Global Class Prototype for Diversity selection}
We have established a global prototype for each class, which serves as the basis for ensuring the diversity of the selected image categories. Given a labeled image $x_{i}^l$, we obtain a set of RoI features $\mathrm{F}_{i}^{gt}$ from the ground-truth bounding boxes generated by the RoI Head of the Student Detector in each training stage:
\begin{equation}
	\label{f_gt}
	\mathrm{F}_{i}^{g t}=\left\{\left(f_{i, j}^{g t}, y_{i, j}^{g t}\right)\right\}
\end{equation}
where $f_{i, j}^{g t}$ denotes the RoI feature of the j-th ground-truth bounding box, $ y_{i, j}^{g t} \in C$ is its class label while $C$ is the set of total instance categories. We obtain a local prototype for each class by calculating the average of the RoI features by class:
\begin{equation}
	\label{local proto}
	v_{k}=\left\{\begin{matrix}
		&\frac{ {\textstyle \sum_{i,j}}f_{i,j}^{gt}\mathbbm{1}\left ( y_{i,j}^{gt}=k \right )}{ {\textstyle \sum_{i,j}}\mathbbm{1}\left ( y_{i,j}^{gt}=k \right )} & \sum_{i} \mathbbm{1}\left ( y_{i,j}^{gt}=k \right )>0  \\
		& \textbf{0}
		&\sum_{i}\mathbbm{1}\left ( y_{i,j}^{gt}=k \right )=0
	\end{matrix}\right.
\end{equation}
where $v_{k}$ is the local prototype of k-th class in $C$, $\textbf{0}$ denotes the zero vector and $\mathbbm{1}\left ( y_{i,j}^{gt}=k \right )$ is defined as follows:
\begin{equation}
	\label{1}
	\mathbbm{1}\left ( y_{i,j}^{gt}=k \right )=\left\{\begin{matrix}
		&1
		&\text { if } y_{i,j}^{gt}=k\\
		&0
		&\text { if } y_{i,j}^{gt} \ne k
	\end{matrix}\right.
\end{equation}

The global class prototype is updated using the EMA algorithm\cite{r-s-meanteacher}, with the local class prototype serving as a reference:
 \begin{equation}
	\label{ema_proto}
	g_{k}=\alpha g_{k}+(1-\alpha)v_{k}
\end{equation}
where $ g_{k}$ is the global prototype of k-th class in $C$, $\alpha$ denotes the hyper-parameter that is typically closed to 1. By applying semi-supervised training on a small initial set of labeled data, we are able to obtain the initial global class prototype, which serves as a foundation for ensuring the diversity and effectiveness of the subsequent active learning process.

For unlabeled images, we only adopt $x^u \in I_{diff}^{u}$ to measure their diversity and update the global class prototype, which is due to the fact that this part of the images already has a certain level of uncertainty and thus is more representative in unlabeled set. First, to make the updating process of the global class prototype smoother, we sort $x_i^u \in I_{diff}^{u}$ from smallest to largest uncertainty values. As described in Section 2.2.1, the weak augmented unlabeled image $T_{w}\left ( x_{j}^{u} \right )$ is first fed into the Teacher network, which subsequently generates a set of proposal bounding boxes $b_{j}$. $b_{j}$ is then input into the RoI head of the Student network to obtain its RoI feature $f_{i,j}^{pgt}$. To measure the similarity between $f_{i,j}^{pgt}$ and $g_{k}$, we adopt cosine similarity as a metric:
 \begin{equation}
	\label{cosine_sim}
	\operatorname{sim}\left(f_{i, j}^{p g t}, g_{k}\right)=\frac{f_{i, j}^{p g t}\left(g_{k}\right)^{T}}{\left\|g_{k}\right\| \cdot\left\|f_{i, j}^{p g t}\right\|}
\end{equation}

For each RoI feature $ f_{i,j}^{pgt}$ , the class $c$ in the global class prototype with the highest similarity score can be identified, denoted as $\max _{k \in N_{c}} \operatorname{sim}\left(f_{i, j}^{p g t}, g_{k}\right)$. If the similarity score falls below a certain threshold $s$, it suggests that the target object might belong to a novel class that is not yet present in the global class prototype. Consequently, the global class prototype must be updated using equations (12) and (14).

To ensure the diversity of selected samples, we need to suppress instances with high similarity to the global class prototype. Therefore, we can calculate the diversity score $S_{div}^{i}$ of the unlabeled image $x_i^u \in I_{diff}^{u}$:
\begin{equation}
	\label{div}
	S_{div}^{i}=1-\frac{1}{n_{b}^{i}} \sum_{j=1}^{n_{b}^{i}} \max _{k \in N_{c}} \operatorname{sim}\left(f_{i, j}^{p g t}, g_{k}\right)
\end{equation}

Higher $S_{div}^{i}$ means more dissimilar the image is to other images, indicating that it may represent a novel class or a new aspect of a known class. Finally, we use \textit{L-p} normalization to combine the two metrics as the final selection score of the unlabeled image $x_i^u$:
\begin{equation}
	\label{L-p}
	S_{sel}^{i}=\sqrt[p]{\left(S_{unc}^{i}\right)^{p}+\left(S_{div }^{i}\right)^{p}}
\end{equation}

\section{Experiment And Analysis}
\subsection{Datasets and Experiments Designation}
To extensively evaluate the proposed framework, two representative and public datasets in remote-sensing images, known as DOTA\cite{dota} and DIOR\cite{dior} were employed in our experiments.
DOTA\cite{dota} contains 2806 aerial images from different sensors and platforms with crowd sourcing and 188,282 instances, covered by 15 common object categories. 
DIOR\cite{dior} consists of 23,463 images and 192,472 instances that are manually labeled with axis-aligned bounding boxes, covering 20 object categories. 
In our experiments, we focus on the task of detection with horizontal bounding boxes (\textbf{HBB} for short). We divide both of the datasets into training set and test set, with a ratio of 2:1.

Similar to the previous works for SSOD\cite{ubteacher}, \cite{activeteacher}, each training set is randomly divided into the labeled set and the unlabeled set. The labeled set is initialized with 5\% images from the training set of DOTA and 2.5\% of DIOR. For evaluation, we adopt mAP (50:95)\cite{map} as the basic comparison metric. In SSOD-AT, we adopt Faster-RCNN with ResNet-50 as the basic detection network. The batch size for training is set to 32, which consists 16 labeled and 16 unlabeled images via random sampling. For the batch size $h$ of active sampling, we selected the 5\% samples from the unlabeled data in DOTA and the 2.5\% in DIOR for human labeling in each iteration. 
\begin{table*}[tb] 
    \caption{Compared with State-of-the-Art Methods on the DOTA Dataset. The Metric is mAP(50:95). “Supervised” Refers to the Model Fed by Labeled Data Only. * is the Origin SSOD Model Fed by Our Active Sampled Data. $\Delta$: AP Gain to the Supervised Performance} %标题
    \centering
    \label{com-DOTA-1}
    \setlength{\tabcolsep}{1.55mm}{
    \begin{tabular}{|c|cc|cc|cc|cc|cc|cc|cc|}
        \hline
        \makecell[c]{\multirow{2}*{Method}} & \multicolumn{14}{c|}{DOTA} \\
        \cline{2-15}
        & L+5\% & $\Delta$ & L+10\% & $\Delta$ & L+15\% & $\Delta$ & L+20\% & $\Delta$ & L+25\% & $\Delta$ & L+35\% & $\Delta$ & L+45\% & $\Delta$ \\
        \hline
        Supervised & 26.62 & +0.00 & 32.48 & +0.00 & 36.71 & +0.00 & 40.84 & +0.00 & 43.83 & +0.00 & 47.41 & +0.00 & 52.16 & +0.00 \\
        \hline
        CALD & 27.13 & +0.51 & 35.46 & +2.98 & 40.01 & +3.30 & 43.09 & +2.25 & 45.47 & +1.64 & 50.83 & +3.42 & 54.42 & +2.26 \\
        \hline
        Unbiased-Teacher & 44.53 & +17.91 & 47.36 & +14.88 & 49.79 & +13.08 & 52.49 & +11.65 & 54.52 & +10.69 & 57.22 & +9.81 & 59.40 & +7.24 \\
        \hline
        ILNet & 44.08 & +17.46 & 46.81 & +14.33 & 48.35 & +11.64 & 51.12 & +10.28 & 51.46 & +7.63 & 54.77 & +7.36 & 56.19 & +4.03 \\
        \hline
        Unbiased-Teacher* & 44.31 & +17.69 & 49.21 & +16.73 & 52.19 & +15.48 & 53.95 & +13.11 & 55.55 & +11.72 & 57.65 & +10.24 & 59.88 & +7.72 \\
        \hline
        ILNet* & 43.82 & +17.20 & 46.97 & +14.49 & 49.23 & +12.52 & 50.64 & +9.80 & 52.78 & +8.95 & 54.61 & +7.20 & 56.55 & +4.39 \\
        \hline
        Active-Teacher & 44.27 & +17.65 & 45.31 & +12.83 & 46.84 & +10.13 & 48.57 & +7.73 & 51.27 & +7.44 & 56.97 & +9.56 & 60.46 & +8.30 \\
        \hline
        SSOD-AT(Random) & 45.43 & +18.81 & 48.75 & +16.27 & 51.04 & +14.33 & 53.37 & +12.53 & 55.67 & +11.84 & 58.34 & +10.93 & 60.41 & +8.25 \\
        \hline
        SSOD-AT(Ours) & \textbf{45.78} & \textbf{+19.16} & \textbf{49.86} & \textbf{+17.38} & \textbf{53.23} & \textbf{+16.52} & \textbf{55.02} & \textbf{+14.18} & \textbf{56.47} & \textbf{+12.64} & \textbf{59.02} & \textbf{+11.61} & \textbf{60.68} & \textbf{+8.35} \\
        \hline  
    \end{tabular}}
\end{table*}
\begin{table*}[tb] 
    \caption{Compared with State-of-the-Art Methods on the DIOR Dataset. The Metric is mAP(50:95). “Supervised” Refers to the Model Fed by Labeled Data Only. * is the Origin SSOD Model Fed by Our Active Sampled Data. $\Delta$: AP Gain to the Supervised Performance} %标题
    \centering
    \label{com-DIOR-1}
    \setlength{\tabcolsep}{1.4mm}{
    \begin{tabular}{|c|cc|cc|cc|cc|cc|cc|cc|}
        \hline
        \makecell[c]{\multirow{2}*{Method}} & \multicolumn{14}{c|}{DIOR} \\
        \cline{2-15}
        & L+5\% & $\Delta$ & L+7.5\% & $\Delta$ & L+10\% & $\Delta$ & L+12.5\% & $\Delta$ & L+15\% & $\Delta$ & L+20\% & $\Delta$ & L+25\% & $\Delta$ \\
        \hline
        Supervised & 22.87 & +0.00 & 25.28 & +0.00 & 27.03 & +0.00 & 29.34 & +0.00 & 30.45 & +0.00 & 32.49 & +0.00 & 34.61 & +0.00 \\
        \hline
        CALD & 24.10 & +1.23 & 26.96 & +1.68 & 28.85 & +1.82 & 30.93 & +1.59 & 32.21 & +1.76 & 34.57 & +2.08 & 36.45 & +1.84 \\
        \hline
        Unbiased-Teacher & 43.25 & +20.38 & 44.48 & +19.20 & 45.65 & +18.62 & 46.86 & +17.52 & 47.20 & +16.75 & 48.24 & +15.75 & 49.03 & +14.42 \\           
        \hline
        ILNet & 43.91 & +21.04 & 44.55 & +19.27 & 45.69 & +18.66 & 46.61 & +17.27 & 47.20 & +16.75 & 47.65 & +15.16 & 48.63 & +14.02 \\                    
        \hline
        Unbiased-Teacher* & 43.99 & +21.12 & 45.22 & +19.94 & 46.05 & +19.02 & 47.07 & +17.73 & 47.53 & +17.08 & 48.36 & +15.87 & 49.05 & +14.44 \\  
        \hline
        ILNet* & 43.77 & +20.90 & 45.07 & +19.79 & 46.06 & +19.03 & 46.90 & +17.56 & 47.38 & +16.93 & 47.88 & +15.39 & 48.61 & +14.00 \\
        \hline
        Active-Teacher & 43.86 & +20.99 & 44.92 & +19.64 & 45.67 & +18.64 & 46.64 & +17.30 & 47.48 & +17.00 & 48.23 & +15.74 & 49.13 & +14.52 \\       
        \hline
        SSOD-AT(Random) & 44.10 & +21.23 & 45.30 & +20.02 & 46.31 & +19.28 & 46.87 & +17.53 & 47.56 & +17.11 & 48.50 & +16.01 & 49.71 & +15.10 \\ 
        \hline
        SSOD-AT(Ours) & \textbf{44.20} & \textbf{+21.33} & \textbf{45.54} & \textbf{+20.26} & \textbf{46.75} & \textbf{+19.72} & \textbf{47.67} & \textbf{+18.33} & \textbf{48.14} & \textbf{+17.69} & \textbf{49.21} & \textbf{+16.72} & \textbf{49.90} & \textbf{+15.29} \\
        \hline  
    \end{tabular}}
\end{table*}
\begin{figure}[tb]
    \centering
    \begin{minipage}[t]{0.49\linewidth}
        \centering
	\includegraphics[width=\textwidth]{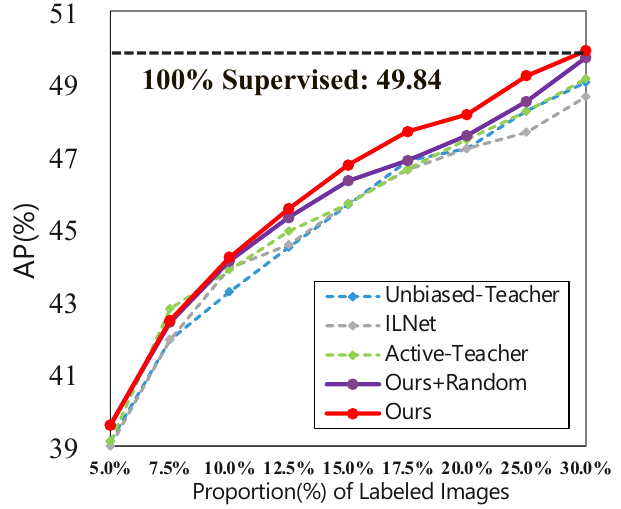}
        \centerline{(a)}
    \end{minipage}
    \begin{minipage}[t]{0.49\linewidth}
        \centering
	\includegraphics[width=\textwidth]{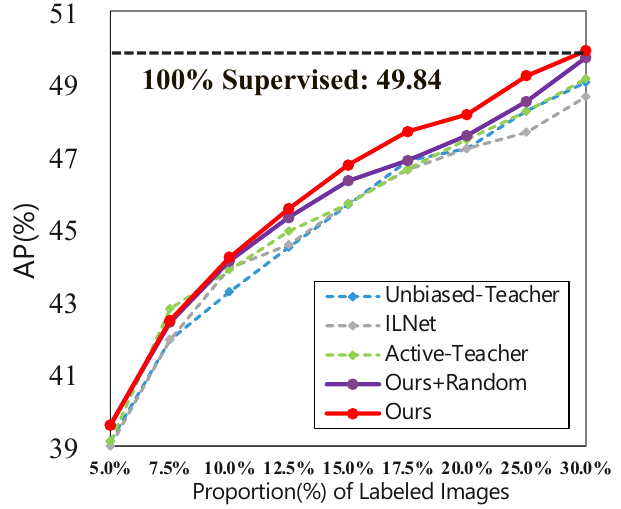}
        \centerline{(b)}
    \end{minipage}
    \caption{\centering{Detection results of the different algorithms on the two remote-sensing datasets. (a) DOTA. (b) DIOR }}
    \label{fig-results}
\end{figure}
\subsection{Experimental Results and Analysis}

\subsubsection{Compare With State-of-the-Art Methods}
Table \ref{com-DOTA-1} and Table \ref{com-DIOR-1} show the experimental results on DOTA and DIOR respectively. We also visualize the whole active learning procedure in Fig. \ref{fig-results}(a) and Fig. \ref{fig-results}(b) on DOTA and DIOR respectively. From the results, we can observe that the proposed method outperforms the SOTA methods and achieves 1\% improvement at most cases in the whole active learning procedure. For SSOD-AT(Random) with RoICM and random selected images, it consistently outperforms the comparison methods at each labeling proportion, which indicates RoICM is capable for detecting noisy RoIs in RSIs. Meanwhile, we should note that when the Unbiased-Teacher and ILNet are trained with the proposed active learning strategy, i.e., Unbiased-Teacher* and ILNet*, their performances always achieve higher, which further indicates the effectiveness of our active learning strategy. Hence, the proposed SSOD-AT is promising for SSOD in remote sensing images.
\begin{table}[htbp]
    \caption{Ablation Study of Proposed SSOD-AT on DOTA Dataset} %标题
    \centering
    \label{ablation-ud-dota}
    \scalebox{0.9}{
        \begin{tabular}{r|c|c|c|c|c|c}
            \toprule
            \multirow{2}*{Method} & \multirow{2}*{RoICM} & \multicolumn{2}{c}{Sampling Strategy} & \multicolumn{3}{c}{DOTA} \\
            \cmidrule(lr){3-7}
            & & Uncertainty & Diversity & 10.0\% & 15.0\% & 20.0\% \\
            \midrule
            Baseline  & $\times$ & $\times$ & $\times$ & 44.53 & 47.36 & 49.79 \\
            \midrule
            \multirow{4}*{SSOD-AT} & \checkmark & $\times$ & $\times$ & 45.43 & 48.75 & 51.04 \\
            & \checkmark & \checkmark & $\times$ & 45.34 & 49.17 & 51.87 \\
            & \checkmark & $\times$ & \checkmark  & 45.69 & 48.50 & 51.46 \\
            & \checkmark & \checkmark & \checkmark & \textbf{45.87} & \textbf{49.43} & \textbf{52.57} \\
            \bottomrule
        \end{tabular}
    }
\end{table}
\begin{table}[htbp] 
    \caption{Ablation Study of Proposed SSOD-AT on DIOR Dataset} %标题
    \centering
    \label{ablation-ud-dior}
    \scalebox{0.9}{
        \begin{tabular}{r|c|c|c|c|c|c}
            \toprule
            \multirow{2}*{Method} & \multirow{2}*{RoICM} & \multicolumn{2}{c}{Sampling Strategy} & \multicolumn{3}{c}{DIOR} \\
            \cmidrule(lr){3-7}
            & & Uncertainty & Diversity & 10.0\% & 15.0\% & 20.0\% \\
            \midrule
            Baseline  & $\times$ & $\times$ & $\times$ & 43.25 & 45.65 & 47.20 \\
            \midrule
            \multirow{4}*{SSOD-AT}& \checkmark & $\times$ & $\times$ & 44.10 & 46.31 & 47.56 \\
            & \checkmark & \checkmark & $\times$ & 44.15 & 46.60 & 47.88 \\
            & \checkmark & $\times$ & \checkmark  & \textbf{44.28} & 46.37 & 47.82 \\
            & \checkmark & \checkmark & \checkmark & 44.20 & \textbf{46.75} & \textbf{48.14} \\
            \bottomrule
        \end{tabular} 
    }   
\end{table}
\subsubsection{Ablation Study}
We show the variants of SSOD-AT with its important components, e.g. \textit{RoICM} and \textit{Sampling Strategy}, on DOTA and DIOR datasets in Table \ref{ablation-ud-dota} and \ref{ablation-ud-dior}. In both tables, the first row represents a baseline teacher-student SSOD method(Unbiased-Teacher). The second row adds our novel RoICM module into the baseline with training samples randomly selected from datasets. The last three rows shows the effectiveness of the uncertainty and diversity modules in our sample strategy. It can be seen that the average accuracy of SSOD-AT is improved with RoICM and the performance is degrading when uncertainty or diversity strategy is removed. This demonstrates that each component is essential for the proposed method.
\subsubsection{Visualization Analysis}
\begin{figure}[t] \centering
    \includegraphics[width=0.49\textwidth]{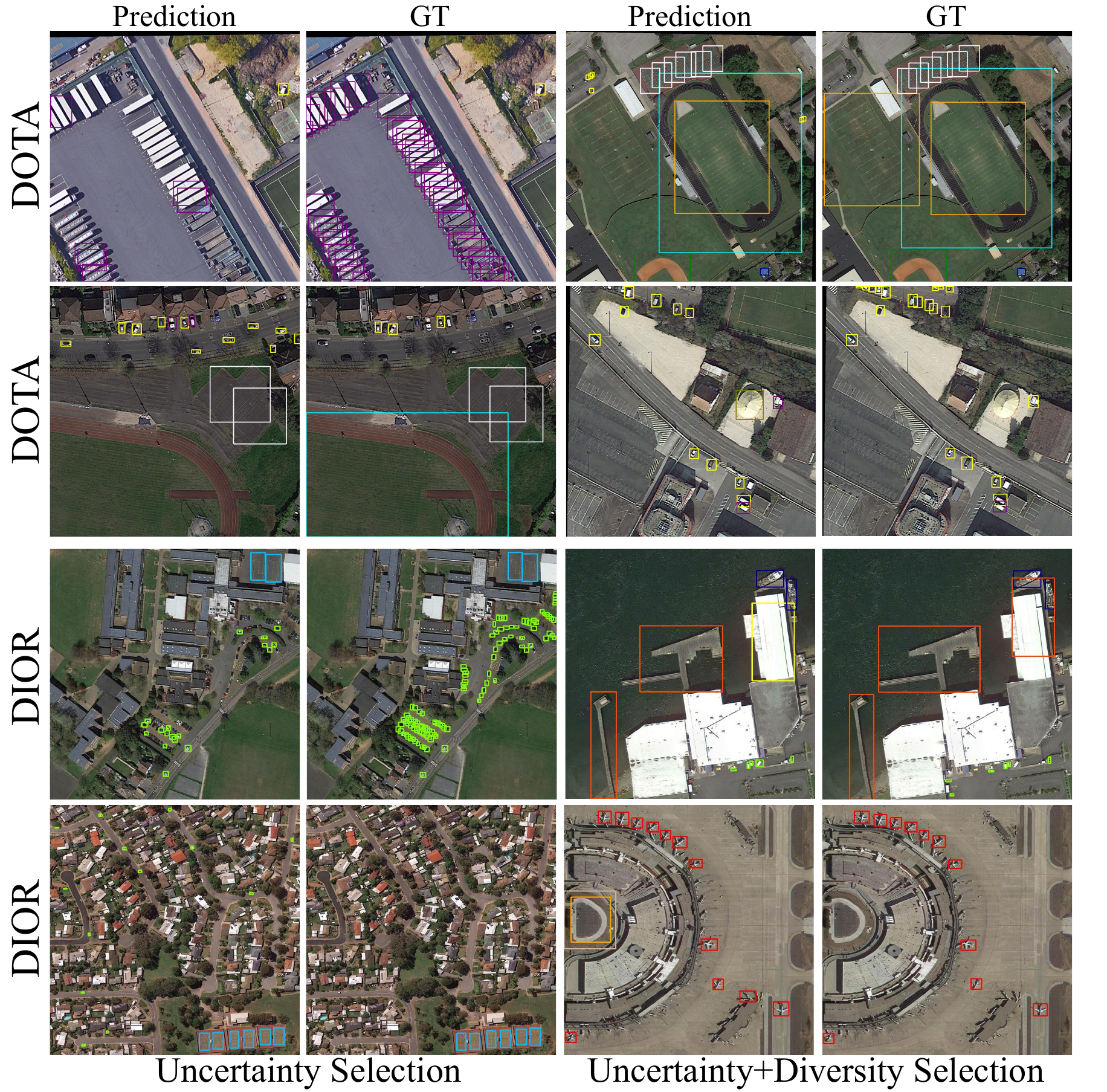}
    \caption{Visualization of the images with top rank selection priority with different active sampling strategies. The Prediction columns denotes the pseudo-labels predicted by teacher network with 30\%(DOTA) and 20\%(DIOR) labeled proportions, while the GT columns refers to the corresponding ground-truths.}
    \label{visual_abla}
\end{figure}
In Fig. \ref{visual_abla}, we visualize the examples selected by active sampling strategies when 30\% images are labeled in DOTA and 20\% images are labeled in DIOR. It is observable that the uncertainty selection usually selects images with objects that are difficult to detect(e.g., small and occluded objects). Comparing Prediction and GT, we can deserve that a large number of objects are missing since the detector is highly uncertain about these images. However, the samples selected by uncertainty are greatly susceptible to be category imbalanced (e.g., a significant fraction of images containing vehicles), while the addition of the diversity selection allows images containing more categories.
% \subsubsection{Visualization of Effects of Proposed RoICM and Sampling Strategies}
% We further visualize the pseudo-labels of SSOD-AT with and without RoICM and active sampling with different training steps and label proportions on DOTA and DIOR datasets. As shown in Fig\ref{visual-effect}, although there is still an obvious gap between the qualities of the pseudo-labels and the ground-truth ones, with the assist of RoICM and our active sampling strategies, SSOD-AT still achieves the desired effects. RoICM can help the detector successfully filter a great part of unreliable pseudo-labels. Besides, SSOD-AT is also able to detect more small objects in image since the incorporation of our AL query criterion.
\section{Conclusion}
We propose a novel method for semi-supervised object detection with active learning for remote sensing images. The proposed method integrates the RoI Comparison Module(RoICM) and Category Prototype to ensure the reliability of pseudo-labels generated by the Teacher Detector and to effectively select the most informative images for expert labeling. Our proposed method is extensively evaluated on two remote sensing datasets, and it consistently outperforms state-of-the-art methods.

\bibliographystyle{IEEEtran} 
 
% \bibliography{AL-ssod}

\begin{thebibliography}{10}
\providecommand{\url}[1]{#1}
\csname url@samestyle\endcsname
\providecommand{\newblock}{\relax}
\providecommand{\bibinfo}[2]{#2}
\providecommand{\BIBentrySTDinterwordspacing}{\spaceskip=0pt\relax}
\providecommand{\BIBentryALTinterwordstretchfactor}{4}
\providecommand{\BIBentryALTinterwordspacing}{\spaceskip=\fontdimen2\font plus
\BIBentryALTinterwordstretchfactor\fontdimen3\font minus \fontdimen4\font\relax}
\providecommand{\BIBforeignlanguage}[2]{{%
\expandafter\ifx\csname l@#1\endcsname\relax
\typeout{** WARNING: IEEEtran.bst: No hyphenation pattern has been}%
\typeout{** loaded for the language `#1'. Using the pattern for}%
\typeout{** the default language instead.}%
\else
\language=\csname l@#1\endcsname
\fi
#2}}
\providecommand{\BIBdecl}{\relax}
\BIBdecl

\bibitem{intro-1}
M.~ElMikaty and T.~Stathaki, ``Detection of cars in high-resolution aerial images of complex urban environments,'' \emph{IEEE Transactions on Geoscience and Remote Sensing}, vol.~55, no.~10, pp. 5913--5924, 2017.

\bibitem{intro-2}
G.~Cheng, J.~Han, P.~Zhou, and D.~Xu, ``Learning rotation-invariant and fisher discriminative convolutional neural networks for object detection,'' \emph{IEEE Transactions on Image Processing}, vol.~28, no.~1, pp. 265--278, 2019.

\bibitem{intro-3}
C.~Li, G.~Cheng, G.~Wang, P.~Zhou, and J.~Han, ``Instance-aware distillation for efficient object detection in remote sensing images,'' \emph{IEEE Transactions on Geoscience and Remote Sensing}, vol.~61, pp. 1--11, 2023.

\bibitem{intro-4}
C.~Li, R.~Cong, C.~Guo, H.~Li, C.~Zhang, F.~Zheng, and Y.~Zhao, ``A parallel down-up fusion network for salient object detection in optical remote sensing images,'' \emph{Neurocomputing}, vol. 415, pp. 411--420, 2020.

\bibitem{intro-5}
W.~Ma, N.~Li, H.~Zhu, L.~Jiao, X.~Tang, Y.~Guo, and B.~Hou, ``Feature split--merge--enhancement network for remote sensing object detection,'' \emph{IEEE Transactions on Geoscience and Remote Sensing}, vol.~60, pp. 1--17, 2022.

\bibitem{intro-6}
Y.~Liu, Q.~Li, Y.~Yuan, Q.~Du, and Q.~Wang, ``Abnet: Adaptive balanced network for multiscale object detection in remote sensing imagery,'' \emph{IEEE Transactions on Geoscience and Remote Sensing}, vol.~60, pp. 1--14, 2021.

\bibitem{intro-7}
Z.~Dong, M.~Wang, Y.~Wang, Y.~Zhu, and Z.~Zhang, ``Object detection in high resolution remote sensing imagery based on convolutional neural networks with suitable object scale features,'' \emph{IEEE Transactions on Geoscience and Remote Sensing}, vol.~58, no.~3, pp. 2104--2114, 2019.

\bibitem{intro-8}
X.~Sun, P.~Wang, Z.~Yan \emph{et~al.}, ``Fair1m: A benchmark dataset for fine-grained object recognition in high-resolution remote sensing imagery,'' \emph{ISPRS Journal of Photogrammetry and Remote Sensing}, vol. 184, pp. 116--130, 2022.

\bibitem{intro-9}
D.~Wan, R.~Lu, S.~Wang, S.~Shen, T.~Xu, and X.~Lang, ``Yolo-hr: Improved yolov5 for object detection in high-resolution optical remote sensing images,'' \emph{Remote Sensing}, vol.~15, no.~3, p. 614, 2023.

\bibitem{intro-ssod-aug-1}
E.~D. Cubuk, B.~Zoph, D.~Mane, V.~Vasudevan, and Q.~V. Le, ``Autoaugment: Learning augmentation strategies from data,'' in \emph{Proceedings of the IEEE/CVF conference on computer vision and pattern recognition}, 2019, pp. 113--123.

\bibitem{intro-active-1}
S.~Tong and D.~Koller, ``Support vector machine active learning with applications to text classification,'' \emph{Journal of machine learning research}, vol.~2, no. Nov, pp. 45--66, 2001.

\bibitem{intro-al-3}
Y.~Gal, R.~Islam, and Z.~Ghahramani, ``Deep bayesian active learning with image data,'' in \emph{International conference on machine learning}.\hskip 1em plus 0.5em minus 0.4em\relax PMLR, 2017, pp. 1183--1192.

\bibitem{intro-al-4}
S.~Agarwal, H.~Arora, S.~Anand, and C.~Arora, ``Contextual diversity for active learning,'' in \emph{16th European Conference on Computer Vision}.\hskip 1em plus 0.5em minus 0.4em\relax Springer, 2020, pp. 137--153.

\bibitem{r-a-multi-instance}
T.~Yuan, F.~Wan, M.~Fu, J.~Liu, S.~Xu, X.~Ji, and Q.~Ye, ``Multiple instance active learning for object detection,'' in \emph{Proceedings of the IEEE/CVF Conference on Computer Vision and Pattern Recognition}, 2021, pp. 5330--5339.

\bibitem{r-a-cald}
W.~Yu, S.~Zhu, T.~Yang, and C.~Chen, ``Consistency-based active learning for object detection,'' in \emph{Proceedings of the IEEE/CVF Conference on Computer Vision and Pattern Recognition}, 2022, pp. 3951--3960.

\bibitem{mol}
G.~Wang, X.~Zhang, Z.~Peng, X.~Jia, X.~Tang, and L.~Jiao, ``Mol: Towards accurate weakly supervised remote sensing object detection via multi-view noisy learning,'' \emph{ISPRS Journal of Photogrammetry and Remote Sensing}, vol. 196, pp. 457--470, 2023\color{black}.

\bibitem{sod}
Y.~Lv, B.~Liu, J.~Zhang, Y.~Dai, A.~Li, and T.~Zhang, ``Semi-supervised active salient object detection,'' \emph{Pattern Recognition}, vol. 123, p. 108364, 2022.

\bibitem{activeteacher}
P.~Mi, J.~Lin, Y.~Zhou, Y.~Shen, G.~Luo, X.~Sun, L.~Cao, R.~Fu, Q.~Xu, and R.~Ji, ``Active teacher for semi-supervised object detection,'' in \emph{Proceedings of the IEEE/CVF Conference on Computer Vision and Pattern Recognition}, 2022, pp. 14\,482--14\,491.

\bibitem{r-s-meanteacher}
A.~Tarvainen and H.~Valpola, ``Mean teachers are better role models: Weight-averaged consistency targets improve semi-supervised deep learning results,'' \emph{Advances in neural information processing systems}, vol.~30, 2017.

\bibitem{dota}
G.-S. Xia, X.~Bai, J.~Ding, Z.~Zhu, S.~Belongie, J.~Luo, M.~Datcu, M.~Pelillo, and L.~Zhang, ``Dota: A large-scale dataset for object detection in aerial images,'' in \emph{Proceedings of the IEEE conference on computer vision and pattern recognition}, 2018, pp. 3974--3983.

\bibitem{dior}
K.~Li, G.~Wan, G.~Cheng, L.~Meng, and J.~Han, ``Object detection in optical remote sensing images: A survey and a new benchmark,'' \emph{ISPRS journal of photogrammetry and remote sensing}, vol. 159, pp. 296--307, 2020.

\bibitem{ubteacher}
Y.-C. Liu, C.-Y. Ma, Z.~He, C.-W. Kuo, K.~Chen, P.~Zhang, B.~Wu, Z.~Kira, and P.~Vajda, ``Unbiased teacher for semi-supervised object detection,'' \emph{nternational Conference on Learning Representations}, 2021.

\bibitem{map}
T.-Y. Lin, M.~Maire, S.~Belongie, J.~Hays, P.~Perona, D.~Ramanan, P.~Doll{\'a}r, and C.~L. Zitnick, ``Microsoft coco: Common objects in context,'' in \emph{13th European Conference on Computer Vision}, 2014, pp. 740--755.

\end{thebibliography}
% Generated by IEEEtran.bst, version: 1.14 (2015/08/26)

\end{document}